\documentclass[runningheads]{llncs}
\usepackage{graphicx}
\usepackage{comment}

\usepackage{amsmath,amssymb} % define this before the line numbering.
\usepackage{color, colortbl}
\usepackage{pifont}
\usepackage{makecell}
\usepackage{floatrow}
\usepackage{multirow}
\usepackage{microtype}
\usepackage{subfigure}
\usepackage{booktabs} % for professional tables
\usepackage{pbox}
\usepackage{multirow}
\usepackage{pdfpages}

\usepackage{multicol}
\usepackage{makecell}
\usepackage{multirow}
\floatstyle{plaintop}
\restylefloat{table}
\usepackage{xspace}
\makeatletter
\DeclareRobustCommand\onedot{\futurelet\@let@token\@onedot}
\def\@onedot{\ifx\@let@token.\else.\null\fi\xspace}
\newcommand\blfootnote[1]{%
  \begingroup
  \renewcommand\thefootnote{}\footnote{#1}%
  \addtocounter{footnote}{-1}%
  \endgroup
}

\def\eg{\emph{e.g}\onedot} 
\def\ie{\emph{i.e}\onedot}

\def\etal{\emph{et al}\onedot}
\makeatother
\usepackage{tabularx}
\usepackage{wrapfig}
\usepackage{lipsum}
\usepackage[]{algorithm2e}

\definecolor{NVblue}{rgb}{0.07, 0.12, 0.83}

\usepackage[pagebackref=true,breaklinks=true,colorlinks=true,citecolor=NVblue,bookmarks=false]{hyperref}

\newfloatcommand{capbtabbox}{table}[\captop][\FBwidth]

\definecolor{jnk_m}{rgb}{0.5, 0.5, 0.5}
\definecolor{mint}{rgb}{0.24, 0.71, 0.54}
\definecolor{orangepeel}{rgb}{1.0, 0.62, 0.0}
\definecolor{awesome}{rgb}{1.0, 0.13, 0.32}

\usepackage[width=122mm,left=12mm,paperwidth=146mm,height=193mm,top=12mm,paperheight=217mm]{geometry}

\begin{document}
% \renewcommand\thelinenumber{\color[rgb]{0.2,0.5,0.8}\normalfont\sffamily\scriptsize\arabic{linenumber}\color[rgb]{0,0,0}}
% \renewcommand\makeLineNumber {\hss\thelinenumber\ \hspace{6mm} \rlap{\hskip\textwidth\ \hspace{6.5mm}\thelinenumber}}
% \linenumbers
\pagestyle{headings}
\mainmatter
\def\ECCVSubNumber{2788}  % Insert your submission number here

%\title{Self-Supervised Human Mesh Recovery Via Color-Consistent Vertex Localization}
%\title{Self-Supervised Human Mesh Recovery Via Vertex-Color Consensus}

\title{Appearance Consensus Driven Self-Supervised Human Mesh Recovery}

\author{Jogendra Nath Kundu$^1$* \and
Mugalodi Rakesh$^1$* \and Varun Jampani$^2$ \and \\ Rahul Mysore Venkatesh$^1$ \and R. Venkatesh Babu$^1$}

% \index{Kundu, Jogendra Nath} 
% \index{Rakesh$, Mugalodi}
% \index{Jampani, Varun}
% \index{Venkatesh, Rahul Mysore}
% \index{Babu, R. Venkatesh}

\titlerunning{Appearance Consensus Driven Self-Supervised Human Mesh Recovery} 
\authorrunning{Kundu \etal}

\institute{$^1$Indian Institute of Science, Bangalore  \qquad $^2$Google Research}

\maketitle

%%%%%%%%%%%%%%%%%%%%%%%%%%%%%%%%%%%%%%%%%%%%%%%%%%%%%%%%%%%%%%
\begin{abstract}
We present a self-supervised human mesh recovery framework\blfootnote{* Equal contribution. $|$ Webpage: \url{https://sites.google.com/view/ss-human-mesh}} to infer human pose and shape from monocular images in the absence of any paired supervision. Recent advances have shifted the interest towards directly regressing parameters of a parametric human model by supervising them on large-scale datasets with 2D landmark annotations. This limits the generalizability of such approaches to operate on images from unlabeled wild environments. Acknowledging this we propose a novel appearance consensus driven self-supervised objective. To effectively disentangle the foreground (FG) human we rely on image pairs depicting the same person (consistent FG) in varied pose and background (BG) which are obtained from unlabeled wild videos. The proposed FG appearance consistency objective makes use of a novel, differentiable \textit{Color-recovery} module to obtain vertex colors without the need for any appearance network; via efficient realization of color-picking and reflectional symmetry. We achieve state-of-the-art results on the standard model-based 3D pose estimation benchmarks at comparable supervision levels. Furthermore, the resulting colored mesh prediction opens up the usage of our framework for a variety of appearance-related tasks beyond the pose and shape estimation, thus establishing our superior generalizability.
%\keywords{TODO}
\end{abstract}
%%%%%%%%%%%%%%%%%%%%%%%%%%%%%%%%%%%%%%%%%%%%%%%%%%%%%%%%%%%%%%

\section{Introduction}
Inferring highly deformable 3D human pose and shape from in-the-wild monocular images has been a longstanding goal in the vision community~\cite{hogg1983model}. This is considered as a key step for a wide range of downstream applications such as robot interaction, rehabilitation guidance, animation industry, etc. Being one of the important subtasks, human pose estimation has gained considerable performance improvements in recent years~\cite{sun2018integral,martinez2017simple,rogez2017lcr}, but in a fully-supervised setting. Such approaches heavily rely on large-scale 2D or 3D pose annotations. Following this, the parametric models of human body, such as SCAPE~\cite{anguelov2005scape}, SMPL~\cite{loper2015smpl}, SMPL(-X)~\cite{pavlakos2019expressive,romero2017embodied} lead the way for a full 3D pose and shape estimation. Additionally, to suppress the inherent 2D-to-3D ambiguity, researchers have also utilized auxiliary cues of supervision such as temporal consistency~\cite{arnab2019exploiting,sun2019human}, multi-view image pairs~\cite{rhodin2016general,joo2018total,huang2017towards}, or even alternate sensor data from Kinect~\cite{weiss2011home} or IMUs~\cite{von2017sparse}. However, estimating 3D human pose and shape from a single RGB image without relying on any direct supervision remains a very challenging problem.

Early approaches~\cite{bogo2016keep,Guan2009estimating,LassnerClosing} adopt iterative optimization techniques to fit a parametric human model (\eg SMPL) to a given image observation. These works attempt to iteratively estimate the body pose and shape that best describe the available 2D observation, which is most often the 2D landmark annotations. Though these works usually get good body fits, such approaches are slow and heavily rely on the 2D landmark annotations~\cite{andriluka20142d,johnson2010clustered,kundu2018ispa} or predictions of an off-the-shelf, fully-supervised Image-to-2D pose networks. However, the recent advances in deep learning has shifted the interest towards data-driven regression based methods~\cite{kanazawa2018end,tung2017self}, where a deep network directly regresses parameters of the human model for a given input image~\cite{omran2018neural,pavlakos2018learning,zanfir2018deep} in a single-shot computation. This is a promising direction as the network can utilize the full image information instead of just the sparse landmarks to estimate human body shape and pose. In the absence of datasets having images with 3D pose and shape ground-truth (GT), several recent works leverage a variety of available paired 2D annotations~\cite{texturepose,tan2018indirect} such as 2D landmarks or silhouettes~\cite{pavlakos2018learning}; alongside the unpaired 3D pose samples to instill the 3D pose priors~\cite{kanazawa2018end} (\ie to assure recovery of valid 3D poses). The strong reliance on paired 2D keypoint ground-truth limits the generalization of such approaches when applied to images from an unseen wild environment. Given the transient nature of human fashion, the visual appearance of human attire keeps evolving. This demands such approaches to periodically update their 2D pose dataset in order to retain their functionality.

%%%%%%%%%%%%%%%%%% Figure 1  %%%%%%%%%%%%%%%
\begin{figure*}[t]  %% 0.94
    \centering
    \includegraphics[width=0.99\textwidth]{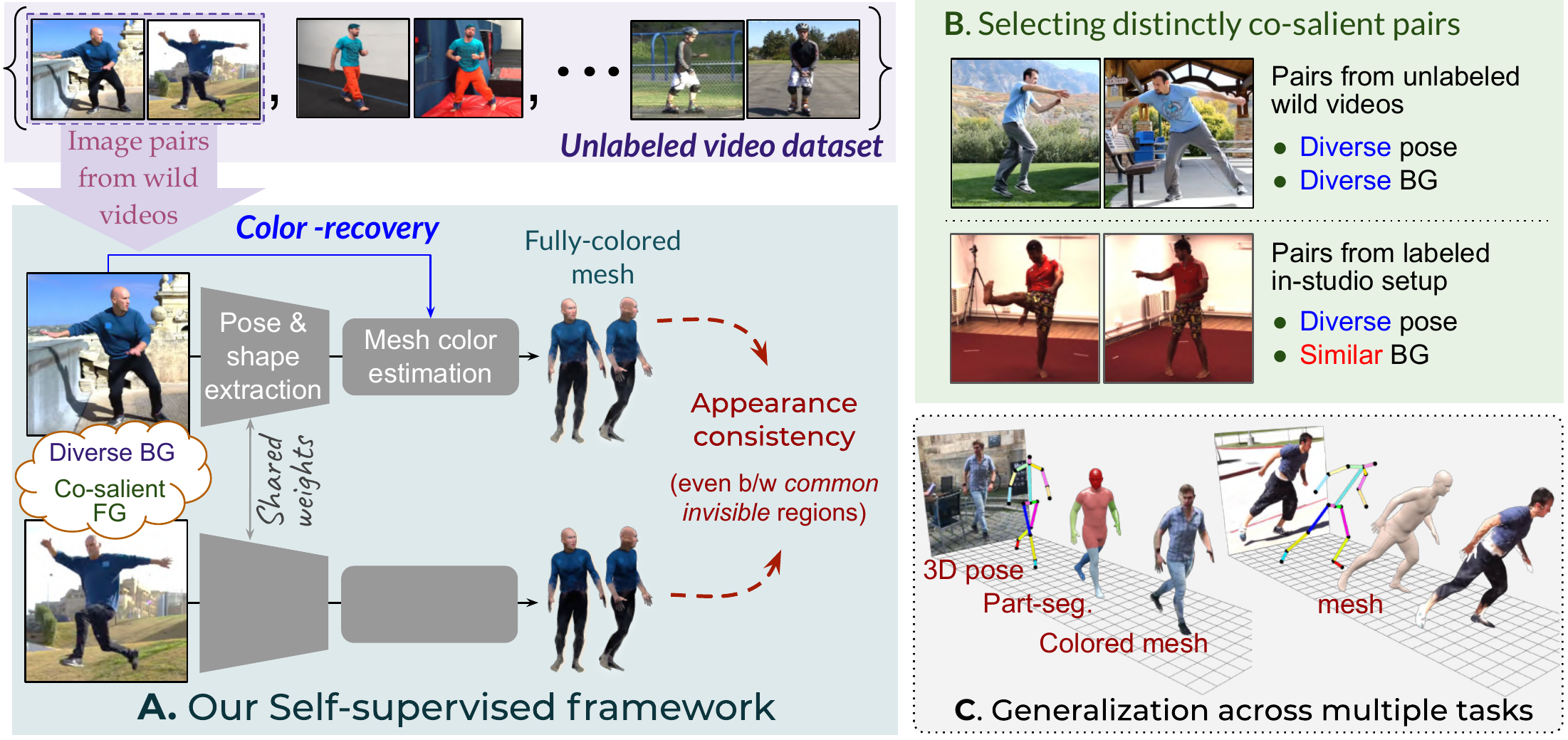} %%\vspace{-2mm}
    \caption{\small Our framework disentangles the co-salient FG human from input image pairs. The resulting colored mesh prediction opens up its usage for a variety of tasks.
    }
    \label{fig_concept}
\end{figure*}

%%%%%%%%%%%%%%%%%%%%%%%%%%%%%%%%%%%%%%%%%%%%%

In this work, the overarching objective is to move away from any kind of paired pose-related supervision for superior generalizability. Our aim is to explore a form of self-supervised objective which can learn both pose and shape from monocular images without accessing any paired GT annotations. We draw motivation from works~\cite{mathieu2016disentangling,rifai2012disentangling,ma2018disentangled,kundu2020self} that aim to disentangle the fundamental factors of variations from a given image. For human-centric images~\cite{kundu2020kinematic}, these factors could be; a) pose, b) foreground (FG) appearance, and c) background (BG) appearance. Here, we leverage the full advantage of incorporating a parametric human model in our framework. Note that, this parametric model not only encapsulates the pose but also segregates the FG region from the BG, which is enabled by projecting the 3D mesh onto the image plane. Thus, the problem boils down to a faithful registration of the 3D mesh onto the image plane or in other words disentanglement of FG from BG. To achieve this disentanglement, we rely on {image pairs} depicting consistent FG appearance but varied 3D poses. Such image pairs can be obtained from videos depicting actions of a single person, which are abundantly available on the internet. %(\eg YouTube). 
Our idea stems from the concept of co-saliency detection~\cite{zhang2016co,hsu2018unsupervised} where the objective is to segment out the common, salient FG from a set of two or more images.  %(see Fig.~\ref{fig_concept}B). 
Surprisingly, this idea works the best for image pairs sampled from wild videos as compared to videos captured in a constrained in-studio setup (static homogeneous background). This is because in wild scenarios, the commonness of FG is distinctly %more
salient in relatively diverse BGs as a result of substantial camera movements (see Fig.~\ref{fig_concept}{\color{red}B}). Thus, in contrast to prior self-supervised approaches that either rely on videos with static BG~\cite{rhodin2018unsupervised} or operate under the assumption of BG commonness between temporally close frames~\cite{jakab2018unsupervised}; our approach is more favorable to learn from wild videos hence better generalizable.

%%%%%%%%%%%%%%%%%%%%%%%%%%%%%%%%%%%%%%%%%%%%%%%%
\begin{wraptable}[9]{r}{7.5cm} %[9]
%\vspace{-6mm}
\caption{%\scriptsize 
Characteristic 
comparison against prior-arts.
} %\vspace{-3mm}
\centering
\label{tab_char}
\setlength{\tabcolsep}{0.4pt}
\resizebox{1.0\linewidth}{!}{
    %\vspace{-4mm}
    \begin{tabular}{l|c|c|c}
    \hline%\noalign{\smallskip}
    \makecell{Model-based\\ methods} & \makecell{2D keypoint\\ supervision} & \makecell{Temporal\\supervision} & \makecell{Colored mesh \\prediction} \\
    \hline\hline
    \cite{kanazawa2018end,kolotouros2019learning,kolotouros2019nal,pavlakos2018learning,omran2018neural} & Yes & No & No \\
    \cite{sun2019human,arnab2019exploiting,kanazawa2019learning} & Yes & Yes & No \\ \hline
    Ours(self-sup.) & \textbf{No} & \textbf{No} & \textbf{Yes} \\
    \hline
\end{tabular} %\vspace{-10mm}
}
\end{wraptable}
%%%%%%%%%%%%%%%%%%%%%%%%%%%%%%%%%%%%%%%%%%%%%%%%

In the proposed framework, we first employ a CNN regressor to obtain the parameters (both pose and shape) of the SMPL model for a given input image. The human mesh model uses these parameters to output the mesh vertex locations. In contrast to the general trend~\cite{alp2018densepose,kanazawa2018learning}, we propose a novel way of inferring mesh texture where the network’s burden to regress vertex color or any sort of appearance representation (such as UV map) is entirely taken away. This is realized via a differentiable \textit{Color-recovery} module which aims to assign color to the mesh vertices via spatial registration of the mesh over the image plane while effectively accounting for the challenges of mesh-vertex visibility like self and inter-part occlusions. To obtain a {fully-colored mesh}, we use a predefined, 4-way symmetry grouping knowledge (front-back and left-right) to propagate the color from camera visible vertices to the non-visible ones in a fully differentiable fashion.

For a given image pair, we pass them through two parallel pathways of our colored mesh prediction framework (see Fig.~\ref{fig_concept}{\color{red}A}). The commonness of FG appearance allows us to impose an appearance consistency loss between the predicted mesh representations. In the absence of any paired supervision, this appearance consistency not only helps us to segregate the common FG human from their respective wild BGs but also discovers the required pose deformation in a fully self-supervised manner. {The proposed reflectional symmetry module brings in a substantial advantage in our self-supervised framework by allowing us to impose appearance consistency even between body parts which are ``\textit{commonly invisible}" in both the images}. Recognizing the unreliability of consistent raw color intensities which can easily be violated as result of illumination changes, we propose a \textit{part-prototype} consistency objective. This aims to match a higher level appearance representation beyond the raw color intensities which is enabled by operating the \textit{Color-recovery} module on convolutional feature maps instead of the raw image. Additionally, to regularize the self-supervised framework, we also impose a shape consistency loss alongside the imposition of 3D pose prior learned from a set of unpaired MoCap samples. Note that at test time, we perform single image inference to estimate 3D human pose and shape. % similar to severa approaches.

In summary, we make the following main contributions:
%%%%%%%%%%%%%%%%%%%%%%%%%%%%%%%%%%%%%%%%%%%%%
%%\vspace{-2mm}
\begin{itemize}
    \item We propose a self-supervised learning technique to perform simultaneous pose and shape estimation which uses image pairs sampled from in-the-wild videos in the absence of any paired supervision.
    \item The proposed \textit{Color-recovery} module  
    completely eliminates the network’s burden to regress any appearance-related representation via efficient realization of color-picking and reflectional symmetry. This best suits our self-supervised framework 
    which relies on FG appearance consistency.
    \item We demonstrate generalizability of our framework to operate on \textit{unseen} wild datasets. 
    We achieve \textit{state-of-the-art} results against the prior model-based pose estimation approaches when tested at comparable supervision levels.
    
\end{itemize}
%%%%%%%%%%%%%%%%%%%%%%%%%%%%%%%%%%%%%%%%%%%%%

\section{Related Work}
\textbf{Vertex-color reconstruction.} In literature, we find different ways to infer textured 3D mesh from a monocular RGB image. %As the most trivial way, 
Certain approaches~\cite{l2019differ,song2017semantic} train a deep network to directly regress 3D features (RGB colors) for individual vertices. In the second kind, a fully convolutional deep network is trained to map the location of each pixel to the corresponding continuous UV-map coordinate parameterization~\cite{alp2018densepose}. %This requires a canonical UV image mapping of the mesh vertices, which is independent of the mesh deformations required for the given input image.
In the third kind, the deep model is trained to directly regress the UV-image~\cite{kanazawa2018learning}. Note that, the spatial structure of the UV image is much different from that of the input image which prevents employing a fully-convolutional network for the same. Recently proposed, Soft-Rasterizer~\cite{liu2019soft} uses a color-selection and color-sampling network whose outputs are processed to obtain the final vertex colors. All the above approaches adopt a learnable way to obtain the mesh color (\ie obtained as neural output). In such cases, the deep network requires substantial training iterations to instill the knowledge of pre-defined UV mapping conventions. We believe this is an additional burden for the network specifically in absence of any auxiliary paired supervisions.

%%\vspace{1.5mm}
%%\noindent 
\textbf{Model-based human mesh estimation.} 
Recently, parametric human models~\cite{anguelov2005scape,loper2015smpl} have been used as the output target for the simultaneous pose and shape estimation task. Such a well-defined mesh model with ordered vertices provides a direct mapping to the corresponding 3D pose and part segments. Both optimization~\cite{bogo2016keep,LassnerClosing,zanfir2018monocular} and regression~\cite{kanazawa2018end,omran2018neural,pavlakos2018learning,zanfir2018deep} based approaches estimate the body pose and shape that best describes the available 2D observations such as 2D keypoints~\cite{kanazawa2018end}, silhouettes~\cite{pavlakos2018learning}, body/part segmentation~\cite{omran2018neural} etc. Due to the lack of datasets having wild images with 3D pose and shape GT, most of the above approaches fully rely on the availability of 2D keypoint annotations~\cite{andriluka20142d,lin2014microsoft} followed by different variants of a 2D reprojection loss~\cite{tan2018indirect,tung2017self} (see Table~\ref{tab_char}). 

%%\noindent 
\textbf{Use of auxiliary supervision.}
In the absence of any shape supervision, certain prior works also leverage full mesh supervision available from synthetically rendered human images~\cite{varol2017learning} or images with fairly successful body fits~\cite{LassnerClosing}. 
Furthermore, multi-view image pairs have also been used for 3D pose~\cite{rhodin2018unsupervised} and shape estimation~\cite{hofmann2009multi,liang2019shape} via enforcing consistency of canonical 3D pose across multiple views. %or the 3D mesh vertices across multiple views. 
Liang~\etal~\cite{liang2019shape} use a multi-stage regressor for multi-view images to further reduce the projection ambiguity in order to obtain a better performance for 3D human body under clothing. To inculcate strong 3D pose prior, Zhou~\etal~\cite{zhou2017towards} makes use of left-right symmetric bone-length constraint for the skeleton based 3D pose estimation task. Further, to assure recovery of valid 3D poses for the model-based pose estimation task, Kanazawa~\etal~\cite{kanazawa2018end} enforce learning based human pose and shape prior via adversarial networks using unpaired sample of plausible 3D pose and shape. With the advent of differentiable renderers~\cite{henderson19ijcv,kato2018renderer} certain methods supervise 3D shape and pose estimation through a textured mesh prediction network to encourage matching of the rendered texture image with the image FG~\cite{kanazawa2018learning}, alongside the 2D keypoint supervision~\cite{texturepose}. %However, in this paper, we aim to solve human mesh estimation problem in the absence of any paired 2D or 3D pose annotations.

%%%%%%%%%%%%%%%%%%%%%%%%%%%%%%%%%%%%%%%%%%%%%%%%%%
%                                         _     
%                                        | |    
%   __ _ _ __  _ __  _ __ ___   __ _  ___| |__  
%  / _` | '_ \| '_ \| '__/ _ \ / _` |/ __| '_ \ 
% | (_| | |_) | |_) | | | (_) | (_| | (__| | | |
%  \__,_| .__/| .__/|_|  \___/ \__,_|\___|_| |_|
%       | |   | |                               
%       |_|   |_|  
%%%%%%%%%%%%%%%%%%%%%%%%%%%%%%%%%%%%%%%%%%%%%%%%%%

%%\vspace{-2mm}
\section{Approach} %%\vspace{-1mm}
We aim to discover the 3D human pose and shape from unlabeled image pairs of consistent FG appearance. During training, we assume access to a % well-defined 
parametric human mesh model to aid our self-supervised paradigm. 
The mesh model provides a low dimensional parametric representation of variations in human shape and pose deformations. However, by design, this model is unaware of the plausibility restrictions of human pose and shape. Thus, it is prone to implausible poses and self-penetrations specifically in the absence of paired 3D supervision~\cite{kanazawa2018end}. Therefore, to constrain the pose predictions, we assume access to a pool of human 3D pose samples to learn a 3D pose prior. %We refer this data as \textit{unpaired} since we do not have access to the corresponding images.

Fig.~\ref{fig_approach} shows an overview %of the architecture employed in
of our training approach.
% the proposed training paradigm. 
For a given image pair, two parallel pathways of shared CNN regressors predict the human shape and pose parameters alongside the required camera settings to segregate the co-salient FG human. Moreover, to realize a colored mesh representation, we develop a differentiable \textit{Color-recovery} module which infers mesh vertex colors directly from the given image without employing any explicit appearance extraction network.

%%\vspace{-2.5mm}
\subsection{Representation and notations}\label{sec_1}
%%\vspace{-1.5mm}
\noindent \textbf{Human mesh model.} We employ the widely used SMPL body model~\cite{loper2015smpl} which parameterizes a triangulated human mesh of $K=6890$ vertices. This model factorizes the mesh deformations into shape $\beta\in\mathbb{R}^{10}$ and pose $\theta\in\mathbb{R}^{3J}$ with $J=23$ skeleton joints~\cite{kanazawa2018end}. %Here, the shape parameters are the first 10 PCA coefficients of the shape space.
We use the first 10 PCA coefficients of the shape space as a compact shape representation inline with~\cite{kanazawa2018end}. And, the pose is parameterized as parent-relative rotations in the axis-angle representation. This differentiable SMPL function outputs mesh vertex locations in a canonical 3D space which is represented as $V\in\mathbb{R}^{K\times 3}=\mathcal{M}(\theta, \beta)$. Here, the corresponding 3D pose (\ie 3D location of $J$ joints) is obtained using a pre-trained linear regressor, \ie $Y\in\mathbb{R}^{J\times 3}=W_pV$ parameterized by $W_p\in\mathbb{R}^{J\times K}$.
RGB color corresponding to the mesh vertices, $V$
%directly from the input image 
is denoted as $C\in \mathbb{R}^{3\times K}=\textit{CRM}(V, I)$, where \textit{CRM} is the \textit{Color-recovery} module. For each vertex id $k$, $C^{(k)}$ stores the corresponding RGB color intensities. As shown in Fig.~\ref{fig_approach}, we use subscripts $a$ and $b$ to associate
the terms with the respective input images, $I_a$ and $I_b$.

%%\vspace{1mm}
%\noindent 
\textbf{Camera model.} We define a weak perspective camera model using a global orientation $R\in\mathbb{R}^{3\times 3}$ in axis-angle representation (3 angle parameters), a translation $t\in\mathbb{R}^{2}$ and a scale $s\in\mathbb{R}$. Given these parameters, the 2D camera space coordinates of the 3D mesh vertices with vertex index $k$ is obtained as $v^{(k)}=\pi(V^{(k)}) = s\Pi(RV^{(k)})+t$; $v^{(k)}\in\mathcal{U}$, 
where $\Pi$ denotes orthographic projection and $\mathcal{U}\subset \mathbb{R}^2$ denotes the space of image coordinates. Similarly, the camera projected 2D joint locations (2D pose) is expressed as $y\in\mathbb{R}^{J\times 2}=\pi(Y)$.

%%\vspace{-2.5mm}
\subsection{Mesh estimation architecture}\label{sec_2}
%%\vspace{-1.5mm}
For a given monocular image, $I$ as input, we first employ a CNN regressor to predict the SMPL parameters (\ie $\theta$ and $\beta$) alongside the camera parameters, $(R, s, t)$. This is followed by the \textit{Color-recovery} module. The prime functionality of this module is to assign color to the 3D mesh vertices, $C^{(k)}$; $k=1,2,...K$ based on the corresponding image space coordinates obtained via camera projection. However, a reliable color assignment requires us to segregate the vertices based on the following two important criteria.

a) \textbf{Non-camera-facing vertices}: First, the camera-facing vertices are %has to be 
separated from the non-camera-facing ones using the mesh vertex normals. Here, the vertex normal is computed as the normalized average of the surface normals of the faces connected to a given vertex. We first transform these normals from the default canonical system to the camera coordinate system. Following this, Z-component of the \textit{camera-space-normals}, $N^{(k)}\in\mathbb{R}$ are used to segregate the non-camera-facing vertices via a \textit{sigmoid} operation, as shown in Fig.~\ref{fig_approach}. 

b) \textbf{Camera-facing, self-occluded vertices}: Note that, $N^{(k)}$ can not be used to select all the %reliable,
camera-visible vertices in presence of inter-part occlusions (see Fig.~\ref{fig_approach}). As, in such scenario, there exist mesh vertices which face the camera but are obscured by other camera-facing vertices which are closer to the camera in 3D. This calls for modeling the relative depth of mesh-vertices as the second criteria to reliably select the vertices which are closer to the camera among all the camera-facing vertices projected to a certain spatial region. To realize this, we utilize \textit{camera-space-depths}, %denoted as 
$Z^{(k)}\in\mathbb{R}$ which stores the Z-component (or depth) of the vertex location in the camera transformed space.

\subsubsection{Color-recovery module.}
\begin{figure*}[t]
    \centering %%0.93
    \includegraphics[width=0.9\textwidth]{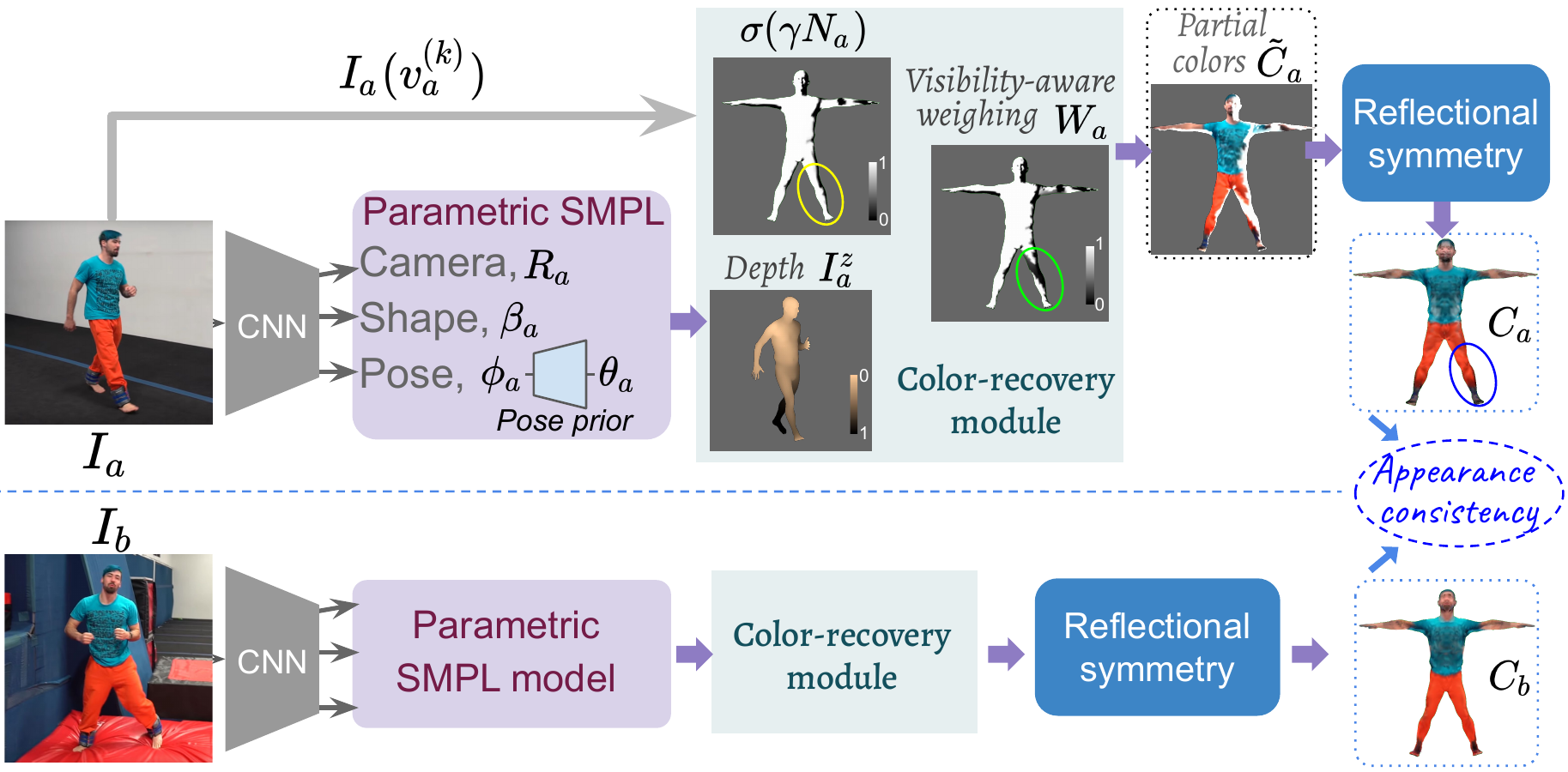}%%\vspace{-3.5mm}
    \caption{The proposed self-supervised %appearance 
    framework makes use of a differentiable \textit{Color-recovery} module to recover the fully colored mesh vertices. \textit{Yellow-circle}: camera-facing vertices does not account for inter-part occlusion. \textit{Green-circle}: $W_a$ accounts for the inter-part occlusion. \textit{Blue-circle}: Fully colored mesh vertices via reflectional symmetry.
    %%\vspace{-3mm}
    }
    \label{fig_approach}
\end{figure*}

In absence of any appearance related features, we plan to realize a spatial depth map using a fast differentiable renderer~\cite{henderson19ijcv} where the camera-space-depth of the mesh vertices, $Z$ is treated as the color intensities for the rendering pipeline. The resultant depth-map is represented as $I^z(u)$, where $u$ spans the space of spatial indices. The % overarching 
general idea is to use this depth-map as a {margin}. 
More concretely, for effective color assignment, one must select the spatially modulated mesh vertices which have the least absolute depth difference with respect to the above defined depth margin. %in spatial proximity. 
To realize this, we compute a depth difference $D^{(k)}$ as $ \vert I^z(v^{(k)}) - Z^{(k)} \vert$, where $I^z(v^{(k)})$ is computed by performing bilinear sampling on $I^z(u)$. In accordance with the above discussion, we formulate a %\textit{occlusion-aware-weighing}
\textit{visibility-aware-weighing} which takes into account both the above mentioned criteria required for an effective mesh vertex selection. %\ie, 

%%\vspace{-4.5mm}
$$
W^{(k)} \in [0,1] = \exp (-\alpha  D^{(k)} )~ \sigma (\gamma N^{(k)}) \text{, where } D^{(k)}= \vert I^z(v^{(k)}) - Z^{(k)} \vert
$$ 
%%\vspace{-0.5mm}
Here, $\exp (-\alpha D^{(k)})$ performs a soft selection by assigning a higher weight value (close to 1) for mesh vertices, $k$ whose \textit{camera-space-depth} $Z^{(k)}$ is in agreement with $I^z(v^{(k)})$ and vice-versa. In the second term, $\sigma$ denotes a sigmoid function with a higher steepness $\gamma$ to reject the non-camera-facing mesh vertices by attributing a low (close to 0) weighing value. %Note that, $W^{(k)}\in [0,1]$. 
Refer Fig.~\ref{fig_approach} for visual illustration.

%%%%%%%%%%%%%%%%%% Figure 6  %%%%%%%%%%%%%%%

%%%%%%%%%%%%%%%%%%%%%%%%%%%%%%%%%%%%%%%%%%%%%

%%\vspace{1mm}
\noindent
\textbf{Intermediate vertex color assignment.}
%\textbf{Obtaining partially colored mesh vertices}
The above defined \textit{visibility-aware-weighing} is employed to realize a primary vertex color assignment. We denote $\tilde{C}\in\mathbb{R}^{3\times K}$ as the intermediate vertex color, where $\tilde{C}^{(k)}$ stores the corresponding RGB color intensities acquired from the given input image $I$. Thus, the primary vertex colors are obtained as, $\tilde{C}^{(k)} = I(v^{(k)}) ~ (2W^{(k)}-1)$, where $I(v^{(k)})$ stores the RGB color intensities at the spatial coordinates $v^{(k)}$ realized via performing bilinear sampling on the input RGB image $I$. The scaled weighing function $(2W^{(k)}-1)$ assigns negative weight to the vertices having low visibility. This assigns a negative color intensity for the corresponding vertices thereby allowing a distinction between the \textit{less-bright} (near-black) %/\textit{darkish-
colors versus \textit{unassigned} vertices.

% \vspace{1mm}
% \noindent 
% \textbf{b) Vertex color assignment via reflectional symmetry.} 

%%\vspace{-4mm}
\subsubsection{Vertex color assignment via reflectional symmetry.}
Here, the prime objective is to propagate the reliable color intensities from the assigned vertices to the unreliable/unassigned ones. The idea is to use reflectional symmetry as a prior knowledge by accessing a predefined set of reflectional groups. For each group-id $g=1,2,...G$, a set of 4 vertices are identified according to left-right and front-back symmetry which would have the same color property (except the vertices belonging to the head where only left-right symmetry is used). This symmetry knowledge is stored as a multi-hot encoding denoted as $S^{(g)}\in \{0,1\}^K$ which constitutes of four ones indicating vertex members in the symmetry group $g$. All the symmetry groups are combined in a symmetry-encoding matrix represented as $S\in \{0,1\}^{G\times K}$. This multi-hot symmetry group representation helps us to perform a fully-differentiable vertex color assignment for all the vertices including the occluded and non-camera facing ones. 

To realize the final vertex colors $C$, we first estimate a group-color for each group $g$ which is denoted by $\mathcal{C}^{(g)}\in\mathbb{R}^{3} = (S^{(g)} \circ \operatorname{ReLU}(\tilde{C})) / (S^{(g)}\circ \operatorname{ReLU}(2W-1))$. Here, $\circ$ denotes dot product between the $K$-dimensional vectors. The group color can be interpreted as a combination of the intermediate vertex colors weighted by their visibility weighing $W$. This effectively handles the cases when only one or more of the vertices in a group are initially colored (visible). That is,
% two diverse cases, \ie a) 
when visibility is active only for a single vertex among the four vertices in a symmetry set; and when visibility is active for all the 4 vertices in a symmetry set; and also the intermediate cases. Finally, the group color is directly propagated to all the mesh vertices using the following matrix multiplication operation, \ie $C=S^T * \mathcal{C}$, where $\mathcal{C}\in\mathbb{R}^{G\times 3} = [\mathcal{C}^{(1)}, \mathcal{C}^{(2)},...\mathcal{C}^{(G)}]$ (see Suppl for more details).

%%\vspace{-2.5mm}
\subsection{Self-supervised learning objectives}\label{sec_3}
%%\vspace{-1.5mm}
For a given image pair, denoted as $I_a$ and $I_b$ (depicting the same person in diverse pose and BGs), we forward them through two parallel pathways of our colored mesh estimation architecture (see Fig.~\ref{fig_approach}). The commonness of FG appearance allows us to impose an appearance consistency loss between the predicted fully colored mesh representations. 

%%\vspace{1mm}
\textbf{a) Color consistency.} First, we impose the following consistency loss,
%%\vspace{-2.0mm}
$$
\boldsymbol{\mathcal{L}_{\textit{CC}}} = \mathcal{L}_{{C}} + \lambda \mathcal{L}_{\tilde{C}} \text{ , where } \mathcal{L}_C=\Vert C_a - C_b \Vert \text{ and } \mathcal{L}_{\tilde{C}}=\Vert W_a\odot W_b\odot (\tilde{C}_a-\tilde{C}_b) \Vert
$$

Here, $\odot$ denotes element-wise multiplication. Note that, $\mathcal{L}_{\tilde{C}}$ enforces a vertex-color consistency on the co-visible mesh vertices (computed as $(W_a\odot W_b)$), \ie the vertices which are visible in both the mesh representations obtained from the image pair, $(I_a, I_b)$. 
However, $\mathcal{L}_{C}$ enforces full vertex color consistency. Here, $\mathcal{L}_{\textit{CC}}$ combines both of the losses thereby providing a higher weightage to the co-visible vertex colors as compared to the approximate full color representation, considering the 
% unreliability 
approximate nature of the symmetry assumption.

%%\vspace{1mm}
\textbf{b) Part-prototype consistency.} The proposed \textit{Color-recovery} module can also be applied on the convolutional feature maps. For a given vertex $k$ and a convolutional feature map $H\in\mathbb{R}^{\tilde{w}\times \tilde{h}\times \tilde{d}}$, we sample $\mathcal{H}^{(k)}\in\mathbb{R}^{\tilde{d}} = H(v^{(k)})$. Note that, we  define a fixed 
%have access to a predefined  (14-parts)
vertex to part-segmentation mapping represented as $Q^{(l)}$, which stores a set of vertex indices for each part $l=1,2,...L$. Now, one can use the vertex visibility weighing $W^{(k)}$ to obtain a prototype appearance feature for each body-part $l$, which is computed as; 
$\mathcal{F}^{(l)} = (\Sigma_{k\in Q^{(l)}} W^{(k)}\mathcal{H}^{(k)}) /( \Sigma_{k\in Q^{(l)}} W^{(k)})$
Following this, we enforce a prototype consistency loss between the image pairs as $\boldsymbol{\mathcal{L}_P} = \Sigma_l \Vert \mathcal{F}_a^{(l)} - \mathcal{F}_b^{(l)} \Vert/ L$. Note that, the prototype feature computation is inherently aware of the inter-part occlusions as a result of incorporating the visibility weighing $W^{(k)}$. As compared to enforcing vertex-color consistency, $\mathcal{L}_{\textit{CC}}$ (\ie the raw color intensities), the part-prototype consistency aims to match a higher-level semantic abstraction (\eg checkered regular patterns %in apparel 
versus just plain {individual colors}) of the part appearances extracted from the image pairs. This also helps us to overcome the unreliability of raw vertex colors which could arise due to illumination differences. 
Motivated by the perceptual loss idea~\cite{johnson2016perceptual}, we obtain $H_a$ and $H_b$
as the \textit{Conv2-1} features corresponding to $I_a$ and $I_b$ from an ImageNet trained (frozen) VGG-16 network~\cite{simonyan2014very}.

%%\vspace{1mm}
\textbf{c) Shape-consistency.} We also enforce a shape consistency loss between the shape parameters obtained from the image pair, \ie $\boldsymbol{\mathcal{L}_\beta} = \vert \beta_a - \beta_b \vert$. Almost all the prior %human mesh estimation 
works~\cite{kanazawa2018end,pavlakos2018learning,texturepose} utilize an \textit{unpaired} human shape dataset to enforce plausibility of the shape predictions via adversarial prior. However, in the proposed self-supervised framework we do not access any human shape dataset. To regularize the shape parameters during the initial training iterations we enforce a loss on shape predictions with respect to a fixed mean shape as a regularization. However, after gaining a decent mesh estimation performance we gradually reduce weightage of this loss by allowing shape variations beyond the mean shape driven by the proposed appearance and shape consistency objectives.

%%\vspace{1mm}
\textbf{d) Enforcing validity of pose predictions.}
Additionally, to assure validity of the predicted pose parameters we train an adversarial auto-encoder~\cite{makhzani2015adversarial} to realize a continuous human pose manifold~\cite{kundu2019bihmp,kundu2019unsupervised} mapped from a latent pose representation, $\phi\in [-1,1]^{32}$. This is trained using an unpaired 3D human pose dataset. The frozen pose decoder obtained from this generative framework is directly employed as a module, with instilled human 3D pose prior. More concretely, a \textit{tanh} non-linearity on the pose-prediction head of the CNN regressor (inline with the latent pose $\phi$) followed by the frozen pose decoder prevents implausible pose predictions during our self-supervised training. In contrast to enforcing an adversarial pose prior objective~\cite{kanazawa2018end,texturepose}, the proposed setup greatly simplifies our training procedure (devoid of discriminator training).

In absence of paired supervision, parameters of the shared CNN regressor is trained by directly enforcing the above consistency losses, \ie $\mathcal{L}_{\textit{CC}}$, $\mathcal{L}_P$, and $\mathcal{L}_\beta$.

%%\vspace{-3mm}
\section{Experiments}%%\vspace{-1.5mm}
We perform thorough experimental analysis to demonstrate the generalizability of our framework across several datasets on a variety of tasks.

%%\vspace{1mm}
\textbf{Implementation details.} We use Resnet-50~\cite{he2016identity} initialized from ImageNet as the base CNN network. The average pooled last layer features are forwarded through a series of fully-connected layers to regress the pose (latent pose encoding $\phi$), shape and camera parameters. Note that, the series of differentiable operations post the CNN regressor do not include any trainable parameters even to estimate the vertex colors. During training, we optimize individual loss terms at alternate training iteration using Adam optimizer~\cite{kingma2014adam}. %We use $\lambda=10$ in the color consistency loss, $\mathcal{L}_\textit{CC}$. 
We enforce prediction of the mean shape for initial 100k training iterations. We also impose a silhouette loss on the predicted human mesh with respect to a pseudo silhouette ground-truth obtained either by using an unsupervised saliency detection method~\cite{zhu2014saliency} or by using a background estimate as favourable for static camera scenarios~\cite{rhodin2018unsupervised}.

%%%%%%%%%%%%%%%%%% Figure 6  %%%%%%%%%%%%%%%
\begin{figure*}[t]
    \centering %%0.98, 0.86
    \includegraphics[width=0.92\textwidth]{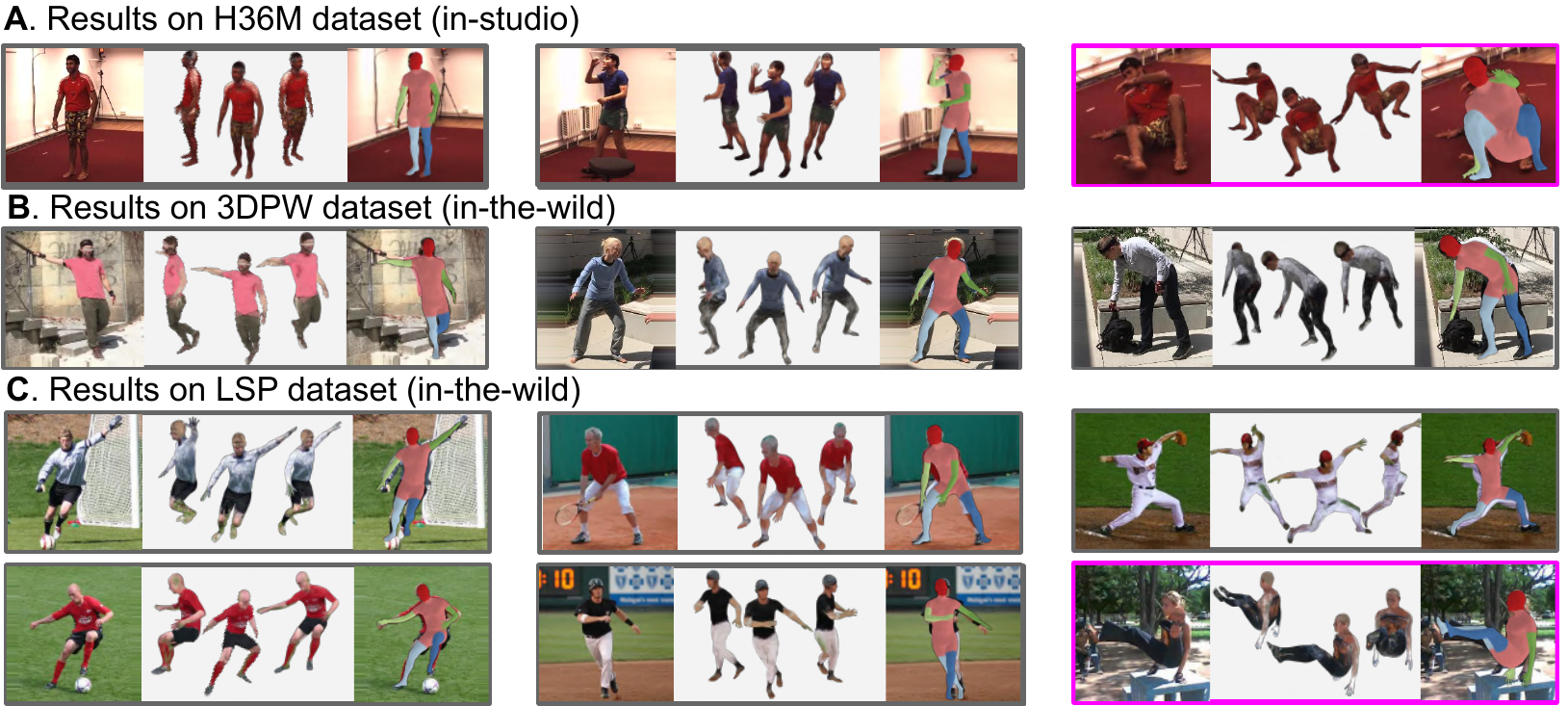}%%\vspace{-2.5mm}
    \caption{Qualitative results. In each panel, 1st column depicts the input image, 2nd column depicts our colored mesh prediction, and 3rd column shows the model-based part segments. Our model fails (in magenta) in presence of complex inter-part occlusions.
    }
    \label{fig_view_syn}
\end{figure*}

%%%%%%%%%%%%%%%%%%%%%%%%%%%%%%%%%%%%%%%%%%%%%

%%\vspace{1mm}
\textbf{Datasets.} 
We sample image pairs with diverse BG (pairs with large \textit{L2} distance) from the following standard datasets, \ie Human3.6M~\cite{ionescu2013human3}, MPII~\cite{andriluka20142d}, MPI-INF-3DHP~\cite{mehta2017monocular} and an in-house collection of wild YouTube videos. In contrast to the in-studio datasets with hardly any camera movement implying static BG~\cite{ionescu2013human3}, the videos collected from YouTube have diverse camera movements (\eg Parkour and Free-running videos). We prune the raw video samples using a %n off-the-shelf 
person-detector~\cite{ren2015faster} to obtain reliable human-centric crops as required for the mesh estimation pipeline (see Suppl). The unpaired 3D pose dataset required to train the 3D pose prior is {obtained from} CMU-MoCap (also used in MoSh~\cite{loper2014mosh}). %(3D MoCap markers).

\textbf{a) Human3.6M} 
This is a widely used dataset consisting of paired image with 3D pose annotations of actors imitating various day-to-day tasks in a controlled in-studio environment. Adhering to well established standards~\cite{kanazawa2018end} we consider subjects S1, S6, S7, S8 for training, S5 for validation and S9, S11 for evaluation, in both Protocol-1 \cite{rhodin2018unsupervised,rhodin2018learning} and Protocol-2 \cite{kanazawa2018end}. %%This dataset constitutes videos captured in a calibrated, time-syncronized 4-view camera setup which also allows us to easily undertake the various ablative studies (such as multi-view supervision), as mentioned in the following subsection. 

\textbf{b) LSP} 
A standard 2D pose dataset consisting of wild athletic actions. We access the LSP test-set with silhouette and part segment annotations as given by Lassner~\etal~\cite{LassnerClosing}.  
In absence of any standard shape evaluation dataset, segmentation results are considered as a proxy for the shape fitting performance~\cite{kanazawa2018end,kolotouros2019nal}.

\textbf{c) 3DPW} 
We also evaluate on the 3D Poses in the Wild dataset~\cite{von2018recovering}. We do not train on 3DPW and use it only to evaluate our cross-dataset generalizability~\cite{kundu2020unsupervised}. We compute the mean per joint position error (MPJPE)~\cite{ionescu2013human3}, both before and after rigid alignment. Rigid alignment is done via Procrustes Analysis~\cite{gower1975generalized}. MPJPE computed post Procrustes alignment is denoted by PA-MPJPE.

%we denote MPJPE computed post alignment step as PA-MPJPE.

%%%%%%%%%%%%%%%%%% Figure 7  %%%%%%%%%%%%%%%
\begin{figure*}[t]
    \centering %%0.98, 0.85
    \includegraphics[width=0.99\textwidth]{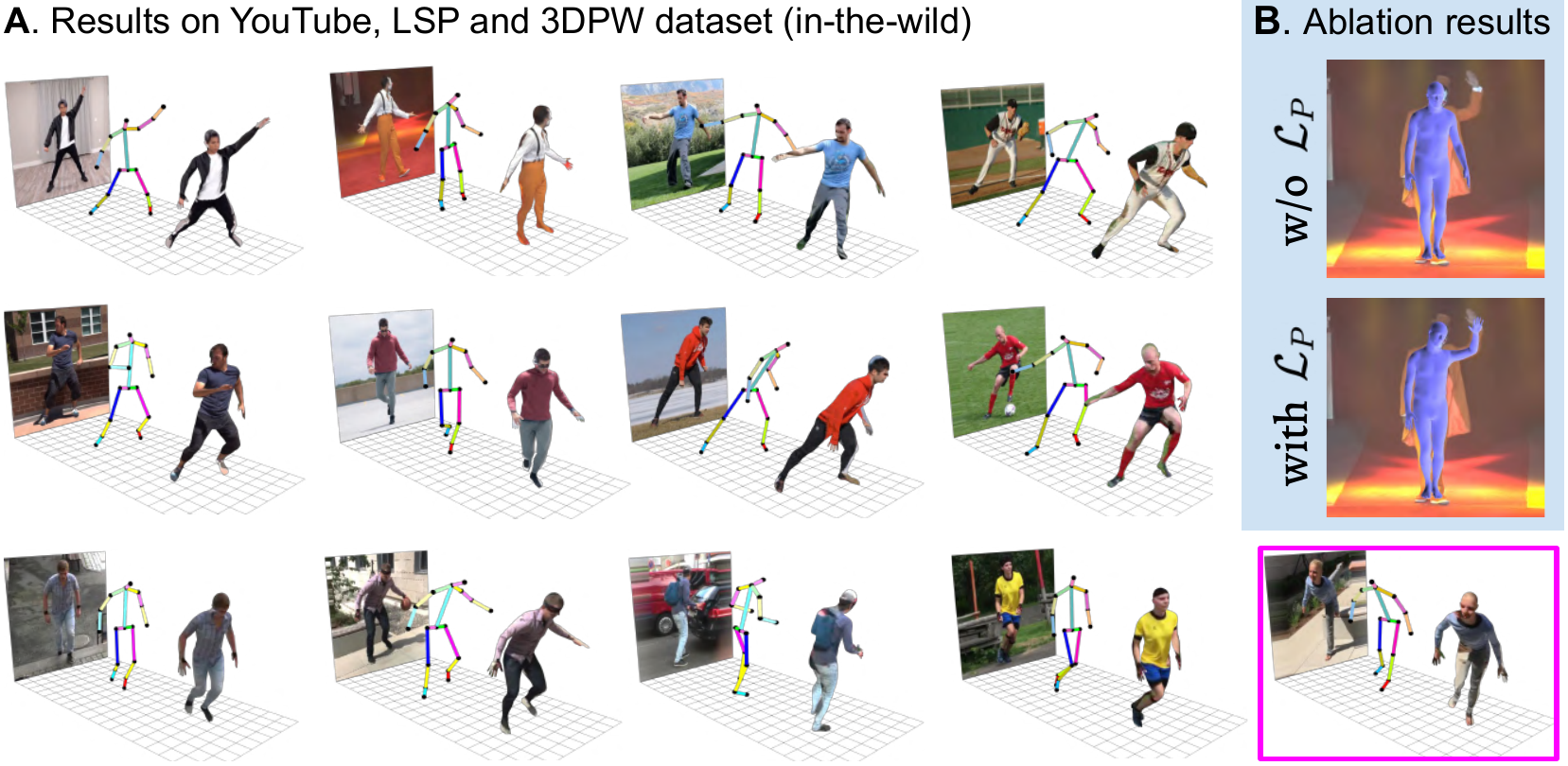}%%\vspace{-1mm}
    \caption{\textbf{A.} Qualitative results on single image colored human mesh recovery. The model fails in presence of complex inter-limb occlusions (in magenta box). \textbf{B.} Qualitative analysis demonstrating importance of incorporating $\mathcal{L}_P$ to extract relevant part-semantics.
    %%\vspace{-5mm}
    }
    \label{fig_box}
\end{figure*}

%%%%%%%%%%%%%%%%%%%%%%%%%%%%%%%%%%%%%%%%%%%%%

%%\vspace{-3mm}
\subsection{Ablative study} %%\vspace{-1.5mm}
To analyze effectiveness of individual self-supervised consistency objectives, we perform ablations by removing certain losses as shown in Table~\ref{tab_ablation}. First, we train \textit{Baseline-1} by enforcing $\mathcal{L}_C$ and $\mathcal{L}_\beta$. Following this, in \textit{Baseline-2} we enforce $\mathcal{L}_{\textit{CC}}$ by incorporating $\mathcal{L}_{\tilde{C}}$ which further penalizes color inconsistency between the vertices which are commonly visible in both the mesh representations. This results in marginal improvement of performance. %as shown Table~\ref{tab_ablation}. 
Moving forward, we recognize a clear limitation in our assumption of FG color consistency (raw RGB intensities) which can easily be violated by illumination differences. Further, the assumption of left-right and front-back symmetry in apparel color can also be violated specifically for {asymmetric upper body apparel}. As a solution, the proposed part-prototype consistency objective, $\mathcal{L}_P$ tries to match a higher level appearance representation beyond just raw color intensities (see Fig.~\ref{fig_box}{\color{red}B}), thus resulting in a significant performance gain (\textit{Ours(unsup)} in Table~\ref{tab_ablation}). Note that, $\mathcal{L}_P$ is possible as a consequence of the proposed differentiable \textit{Color-recovery} module. 

%%%%%%%%%%%%%%%%%%%%%%%%%%%%%%%%%%%%%%
\begin{figure*}[t]
\CenterFloatBoxes
\begin{floatrow}

%%%%%%%%%%%%%%% tab_ablation %%%%%%%%%
\capbtabbox{%
\setlength{\tabcolsep}{1.8pt}
\resizebox{0.99\linewidth}{!}{
    \caption{Ablative study (on Human3.6M) to analyze importance of self-supervised objectives (first 3 rows), and results at varied degree of paired supervision (last 3 rows). P1 and P2 denote MPJPE and PA-MPJPE in Protocol-1 and Protocol-2 respectively.
    %%\vspace{-0.9mm}
    } 
    \label{tab_ablation}
    \begin{tabular}{l|ccc}
    %\hline%\noalign{\smallskip} 
    \toprule
    Methods & P1($\downarrow$) && P2($\downarrow$) \\
    \midrule\midrule
    %\hline\hline
    \textit{Baseline-1}; ($\mathcal{L}_C+\mathcal{L}_\beta$) & 127.1 && 101.2 \\
    \textit{Baseline-2}; ($\mathcal{L}_{\textit{CC}}+\mathcal{L}_\beta$) & 119.6 && 97.4 \\
    \textit{Ours(unsup.)}; ($\mathcal{L}_{\textit{CC}}$+$\mathcal{L}_\beta$+$\mathcal{L}_p$) & \textbf{110.8} && \textbf{90.5} \\ 
    \midrule
    \textit{Ours(multi-view-sup)} & 102.1 && 74.1 \\
    \textit{Ours(weakly-sup)} & 86.4 && 58.2 \\
    \textit{Ours(semi-sup)} & \textbf{73.8} && \textbf{48.1} \\
    %\hline
    \bottomrule
\end{tabular}}}

%%%%%%%%%%%%%%%% tab_3dpw %%%%%%%%%
\capbtabbox{%
\setlength{\tabcolsep}{1.1pt}
\resizebox{0.99\linewidth}{!}{
    \caption{Evaluation on wild 3DPW dataset in a \textit{fully-unseen} setting. Note that, in contrast to Temporal-HMR \cite{kanazawa2019learning} we do not use any temporal supervision. 
    Methods in first 5 rows use equivalent 2D and 3D pose supervision, thus directly comparable. 
    %%\vspace{-5mm}
    } 
    \label{tab_3dpw}
    \begin{tabular}{l|ccc}
    \toprule%\noalign{\smallskip}
    Methods & MPJPE($\downarrow$) && PA-MPJPE($\downarrow$) \\
    \midrule\midrule
    %\multicolumn{4}{c}{Full 2D pose sup. + 3D pose sup. on H3.6M}\\
    Martinez~\etal~\cite{martinez2017simple} & - && 157.0 \\
    SMPLify~\cite{bogo2016keep} & 199.2 && 106.1 \\
    TP-Net~\cite{Dabral_2018_ECCV} & 163.7 && 92.3 \\
    Temporal-HMR~\cite{kanazawa2019learning} & 127.1 && 80.1 \\
    \textit{Ours(semi-sup)} & \textbf{125.8} && \textbf{78.2} \\ 
    \midrule
    \textit{Ours(weakly-sup)} & 153.4 && 89.8 \\
    \textit{Ours(unsup)} & 187.1 && 102.7 \\ 
    \bottomrule
\end{tabular}}}
{%
}
\end{floatrow}
%%\vspace{-2mm}
\end{figure*}
%%%%%%%%%%%%%%%%%%%%%%%%%%%%%%%%%%%%%%
%%%%%%%%%%%%%%%%%%%%%%%%%%%%%%%%%%%%%%

Further, maintaining a fair comparison ground against the prior weakly supervised approaches, we train 3 variants of the proposed framework by utilizing increasing level of paired supervisions alongside our self-supervised objectives.

%%\vspace{1.5mm}
\noindent
a) \textbf{\textit{Ours(multi-view-sup)}} Under multi-view supervision, we impose additional consistency loss on the canonically aligned (view-invariant) 3D mesh vertices (\ie $\Vert V_a - V_b\Vert$) and the 3D pose (\ie $\Vert Y_a-Y_b \Vert$) for the time synchronized multi-view pairs, $(I_a, I_b)$. Inline with Rhodin~\etal~\cite{rhodin2018unsupervised}, we also use full 3D pose supervision only for S1 while evaluating on the standard Human3.6M dataset. We outperform Rhodin~\etal~\cite{rhodin2018unsupervised} by a significant margin as reported in the %last two rows of
Table~\ref{tab_h36m}. This is beyond the usual trend of weaker performance in non-parametric approaches against the model-based parametric ones. Thus, we attribute this performance gain to the proposed appearance consensus driven self-supervised objectives.

%%\vspace{1.5mm}
\noindent
b) \textbf{\textit{Ours(weakly-sup)}} In this setting, we access image datasets with paired 2D landmark annotations, inline with the supervision setting of prior model-based approaches~\cite{kanazawa2018end}. Alongside the proposed self-supervised objectives, we impose a direct 2D landmark supervision loss (\ie $\Vert y-y_\textit{gt}\Vert$) with respect to the corresponding ground-truths but only on samples from specific datasets, such as LSP, LSP-extended~\cite{johnsonclustered} and MPII~\cite{andriluka20142d}. Certain prior arts, such as HMR~\cite{kanazawa2018end}, use even more images with paired 2D landmark annotations from COCO~\cite{lin2014microsoft}.

%%\vspace{1.5mm}
\noindent
c) \textbf{\textit{Ours(semi-sup)}} In this variant, we access paired 3D pose supervision on the widely used in-studio Human3.6M~\cite{ionescu2013human3} dataset alongside the 2D landmark supervision as used in \textit{Ours(weakly-sup)}. Note that, a better performance on Human3.6M (with limited BG and FG diversity as a result of the in-studio data collection setup) does not translate to the same on wild images as a result of the significant domain gap. As we impose the above supervisions alongside the proposed self-supervised objective on unlabeled wild images, such a training is expected to deliver improved performance by successfully overcoming the domain-shift issue. We evaluate this on the wild 3DPW dataset.

%%\vspace{-3mm}
\subsection{Comparison with the state-of-the-art} %%\vspace{-1.5mm}
\textbf{Evaluation on Human3.6M.} %%\hspace{1mm}
Table~\ref{tab_h36m} shows a comparison of different variants of the proposed framework against the prior-arts which are grouped based on the respective supervision levels. We clearly outperform in all the three groups \ie while accessing comparable a) 3D pose supervision, b) 2D landmark supervision, and c) multi-view supervision. Except Rhodin~\etal~\cite{rhodin2018unsupervised} all the prior works mentioned in Table~\ref{tab_h36m} use parametric human model for the human mesh estimation task. Note the significant performance gain specifically in absence of any 3D pose supervision, \ie for \textit{Ours(weakly-sup)} and \textit{Ours(multi-view-sup)} against the relevant counterparts as reported in the last 4 rows.

%%\vspace{1mm}
\noindent
\textbf{Evaluation on 3DPW.} %%\hspace{1mm}
Table~\ref{tab_3dpw} reports a comparison of different variants of the proposed framework against the prior-arts which use comparable pose supervision as used in \textit{Ours(semi-sup)} (except certain methods, such as HMR~\cite{kanazawa2018end} which use even more supervision on 3D pose from the MPI-INF-3DHP~\cite{mehta2017monocular} dataset). It is worth noting that none of our model variants is trained on the samples from 3DPW dataset (not even in self-supervised paradigm). A better performance in such \textit{unseen} setting highlights our superior cross-dataset generalizability.

%%%%%%%%%%%%%%%% tab_h36m %%%%%%%%%
%%%%%%%%%%%%%%%%%%%%%%%%%%%%%%%%%%%%%%
\begin{figure*}[t]
\CenterFloatBoxes
\begin{floatrow}
\capbtabbox{%
\setlength{\tabcolsep}{3.7pt}
\resizebox{0.99\linewidth}{!}{
    \caption{Evaluation on Human3.6M (Protocol-2). Methods in first 9 rows use equivalent 2D and 3D pose supervision hence are directly comparable. Same analogy applies for the rows 10-11 and 12-13. %%\vspace{-5mm}
    } 
    \label{tab_h36m}
    \begin{tabular}{ll|c}
    \toprule%\noalign{\smallskip}
    No. & Methods & PA-MPJPE($\downarrow$) \\
    \midrule\midrule
    1. & Lassner \etal~\cite{LassnerClosing} & 93.9 \\
    2. & Pavlakos~\etal~\cite{pavlakos2018learning} & 75.9 \\
    3. & Omran~\etal~\cite{omran2018neural} & 59.9 \\
    4. & HMR~\cite{kanazawa2018end} & 56.8 \\
    5. & Temporal HMR~\cite{kanazawa2019learning} & 56.9 \\
    6. & Arnab~\etal~\cite{arnab2019exploiting} & 54.3 \\
    7. & Kolotouros~\etal~\cite{kolotouros2019nal} & 50.1 \\
    8. & TexturePose~\cite{texturepose} & 49.7 \\
    9. & \textit{Ours(semi-sup)} & \textbf{48.1} \\ \midrule
    
    10. & HMR unpaired~\cite{kanazawa2018end} & 66.5 \\
    11. & \textit{Ours(weakly-sup)} & \textbf{58.2} \\ \midrule
    
    12. & Rhodin~\etal~\cite{rhodin2018unsupervised} & 98.2 \\
    13. & \textit{Ours(multi-view-sup)} & \textbf{74.1} \\ 
    \bottomrule
\end{tabular}}}

%%%%%%%%%%%%%%% tab_part_seg %%%%%%%%%
\capbtabbox{%
\setlength{\tabcolsep}{1.1pt}
\resizebox{0.999\linewidth}{!}{
  \caption{Evaluation of FG-BG and 6-part segmentation on LSP test set. It reports accuracy (Acc.) and F1 score values of ours against the prior-arts. %in 4 groups (divided by the horizontal lines). 
  \textbf{First group}: Iterative, \textit{optimization-based} approaches.
  \textbf{Last 3 groups}: \textit{Regression-based} methods grouped based on comparable supervision levels.
    %%\vspace{-1mm}
    }
  \label{tab_part_seg}
    %\centering
    \begin{tabular}{l|cc|cc}
        \toprule
        \multirow{2}{*}{Methods} & \multicolumn{2}{c|}{\textbf{FG-BG Seg.}} &
 		\multicolumn{2}{c}{\textbf{Part Seg.}} \\
 		\cline{2-5} %\\ \hline
 		& Acc.($\uparrow$) & F1($\uparrow$) & Acc.($\uparrow$) & F1($\uparrow$) \\ \midrule\midrule
 		SMPLify \textit{oracle}~\cite{bogo2016keep} &  92.17 & 0.88 & 88.82 & 0.67 \\
 		SMPLify~\cite{bogo2016keep} &  91.89 & 0.88 & 87.71 & 0.64 \\
 		SMPLify on~\cite{pavlakos2018learning} &  92.17 & 0.88 & 88.24 & 0.64 \\
 		Bodynet~\cite{varol2018bodynet} &  92.75 & 0.84 & - & - \\ 
 		\midrule
        HMR~\cite{kanazawa2018end} & 91.67 & 0.87 & 87.12 & 0.60 \\
        Kolotouros~\etal~\cite{kolotouros2019nal} & 91.46 & 0.87 & 88.69 & 0.66 \\
        TexturePose~\cite{texturepose} & 91.82 & 0.87 & 89.00 & 0.67 \\
        \textit{Ours(semi-sup)} & \textbf{91.84} & 0.87 & \textbf{89.08} & 0.67 \\
        \midrule
        HMR unpaired~\cite{kanazawa2018end} & 91.30 & 0.86 & 87.00 & 0.59 \\
       \textit{ Ours(weakly-sup)} & \textbf{91.70} & \textbf{0.87} & \textbf{87.12} & \textbf{0.60} \\ 
        \midrule
        \textit{Ours(unsup)} & 91.46 & 0.86 & 87.26 & 0.64 \\
        \bottomrule
    \end{tabular}
    \label{tab:ablation}}}
{%
}
\end{floatrow}
%%\vspace{-2mm}
\end{figure*}
%%%%%%%%%%%%%%%%%%%%%%%%%%%%%%%%%%%%%%
%%%%%%%%%%%%%%%%%%%%%%%%%%%%%%%%%%%%%%

%%\vspace{1mm}
\noindent
\textbf{Evaluation of part-segmentation.} %%\hspace{1mm} 
We also evaluate our performance on FG-BG segmentation and body part-segmentation tasks which are considered as a proxy to quantify the shape fitting performance. In presence of 2D landmark annotation, iterative model fitting approaches have a clear advantage over the single-shot regressor based approaches as shown in Table~\ref{tab_part_seg}. At comparable supervision, \textit{Ours(semi-sup)} not only outperforms the relevant regression based prior arts but also performs competitive to the iterative model fitting based approaches with a significant advantage on inference time (1 min vs ~0.04 sec). %We also report performance of our self-supervised variant, %such as
Note that, \textit{Ours(unsup)} performs competitive to the prior supervised regression-based approaches%even in absence of any paired supervision,
, thus establishing the importance of FG appearance consistency for accurate shape recovery. %See Suppl for qualitative results.

%%\vspace{-2mm}
\subsection{Qualitative results}
\begin{figure*}[t]
    \centering %%0.98
    \includegraphics[width=0.94\textwidth]{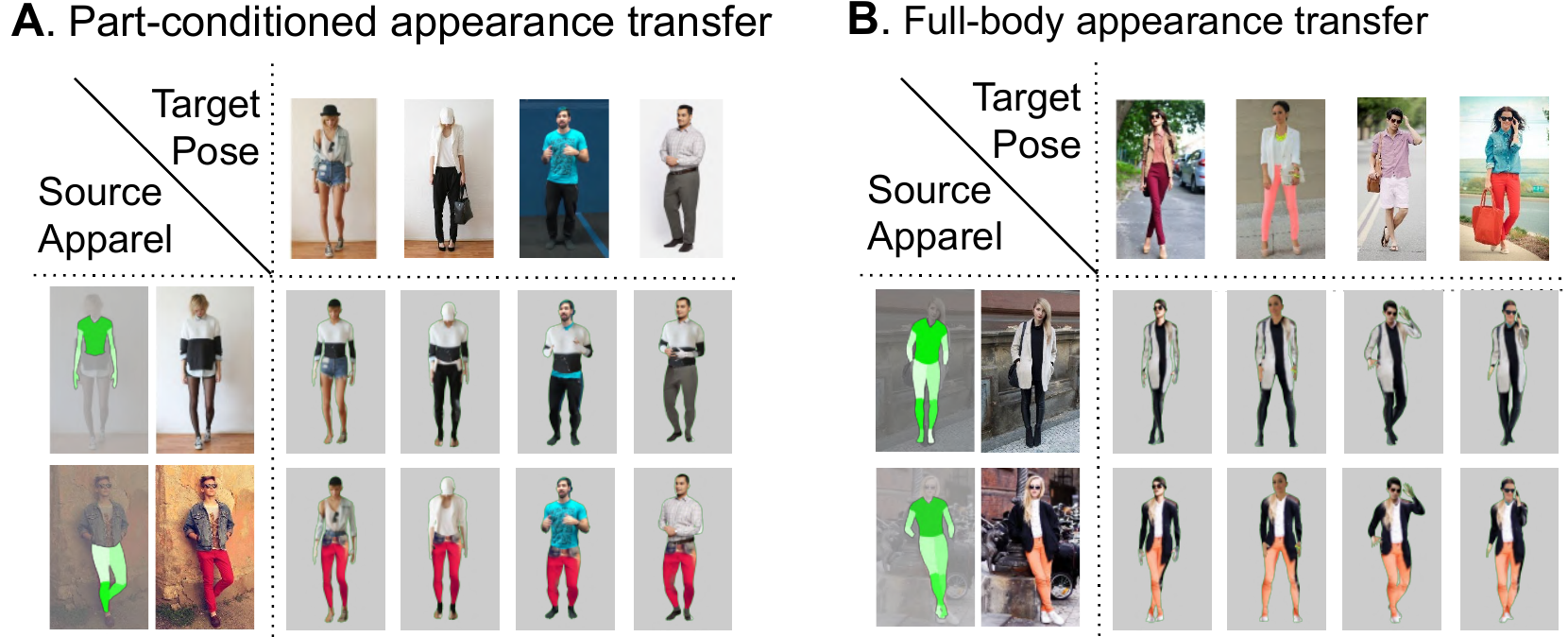}%%\vspace{-2.5mm}
    \caption{Qualitative results on \textbf{A.} {Part-conditioned}, 
    %selective \textit{part-appearance} transfer 
    and \textbf{B.} {Full-body} appearance transfer. This is enabled as a result of our ability to infer the colored mesh representation.
    %%\vspace{-4mm}
    }
    \label{fig_part_syn}
\end{figure*}

The proposed mesh recovery model not only infers pose and shape but also outputs a colored mesh representation as a result of the proposed \textit{reflectional-symmetry} procedure. To evaluate effectiveness of the recovered part appearance we perform 2 different tasks a) part-conditioned appearance transfer, 
%selective part-appearance transfer
and b) full-body appearance transfer as shown in Fig.~\ref{fig_part_syn}. On the top, we show the target images whose pose and shape (network predicted) is combined with part appearances recovered from the source image (only for the highlighted parts) shown on left, to realize a novel synthesized image. Note that, in case of \textit{part-conditioned} appearance transfer, appearance of the non-highlighted parts are taken from the target image shown on the top. For instance, in the first row, the synthesized image depicts upper-body apparel of the person in the source image combined with the lower-body apparel from the target (and in the target image pose). Qualitative results of \textit{Ours(semi-sup)} model on other primary tasks are shown in Fig.~\ref{fig_view_syn} and Fig.~\ref{fig_box} with highlighted failure scenarios (see Suppl).

%%%%%%%%%%%%%%%%%% Figure 5  %%%%%%%%%%%%%%%

%%%%%%%%%%%%%%%%%%%%%%%%%%%%%%%%%%%%%%%%%%%%%

%%\vspace{-2mm}
\section{Conclusion}
We introduce a self-supervised framework for model-based human pose and shape recovery. %Our framework builds on the simple idea of FG appearance consistency. Surprisingly, this idea works the best for image pairs sampled from in-the-wild videos, in contrast to the same from a constrained laboratory setup; as commonness of FG is better highlighted when presented in diverse BGs. 
The proposed appearance consistency not only helps us to segregate the common FG human from their respective wild BGs but also discovers the required pose deformation in a fully self-supervised manner. 
%% The proposed inter-part occlusion aware, \textit{Color-recovery} module assigns color to the highly selective mesh vertices without involving any appearance extraction network. 
However, extending such a framework for human centric images with occlusion by external objects or truncated human visibility, remains to be explored in future.

\noindent
\textbf{Acknowledgements.} We thank Qualcomm Innovation Fellowship India 2020.

\clearpage
% ---- Bibliography ----
%
% BibTeX users should specify bibliography style 'splncs04'.
% References will then be sorted and formatted in the correct style.
%
% \bibliographystyle{splncs04}
% \bibliography{egbib}



\section*{Supplementary material}

# Supplementary Material

## Appearance Consensus Driven
## Self-Supervised Human Mesh Recovery

In this supplementary, we summarize the proposed differentiable colored-mesh recovery procedure followed by additional implementation details and qualitative results. Follow our project page[1] for more details.

The supplementary material is organized as follows:

**Table 1.** A list of notations and their size as used in the main paper.

| Notations | Description | Size |
|---|---|---|
| $I$ | Input image | $224 \times 224 \times 3$ |
| $\theta$ | View invariant SMPL pose | 3J (J=23) |
| $\beta$ | SMPL Shape parameter | 10 |
| $\phi$ | Pose embedding | 32 |
| $V$ | 3D vertex locations | $6890 \times 3$ |
| $C$ | Vertex colors (RGB) | $6890 \times 3$ |
| $v$ | Image-projected vertex locations | $6890 \times 2$ |
| $N$ | Z-component of Camera-space normals | $6890 \times 1$ |
| $Z$ | Camera-space depth of mesh vertices | $6890 \times 1$ |
| $I^z(u)$ | Rendered depth image | $224 \times 224 \times 3$ |
| $W$ | Visibility-aware-weighing | $6890 \times 1$ |
| $\tilde{C}$ | Intermediate Vertex colors (RGB) | $6890 \times 3$ |
| $S$ | Vertex to symmetry group mapping | $(1575+295) \times 6890$ |
| $\mathcal{C}$ | Group color for symmetry groups | $1870 \times 3$ |
| $Q$ | Vertex to part-segmentation mapping | - |
| $H$ | *Conv2-1* output of pre-trained VGG-16 | $112 \times 112 \times 128$ |
| $\mathcal{H}^k$ | Sampled feature from $H$ at $v^{(k)}$ | $\tilde{d} = 128$ |
| $\mathcal{F}$ | Part-prototype appearance feature | 128 |
| $k$ | Index over mesh vertices | K=6890 |
| $g$ | Index over symmetry groups | G=1870 |
| $l$ | Index over body parts | L=14 |
| $a, b$ | Indicating association with inputs $I_a$, $I_b$ | - |

# 1 Differentiable operations in the proposed framework

We propose three completely differentiable modules in order to realize our self-supervised approach namely the color-recovery module, part-prototype module

---

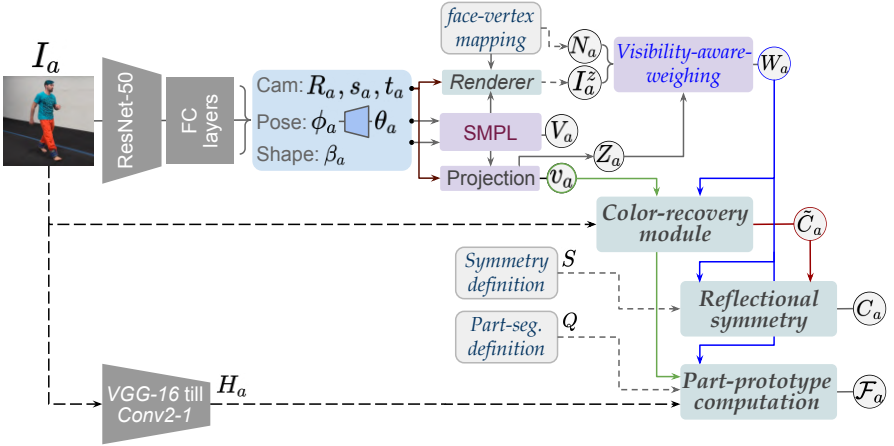

**Fig. 1.** The series of differentiable computations and their interdependence as employed in the proposed self-supervised mesh recovery framework (see Table 1 for the notations).

and the reflectional symmetry module. See Fig. 1 for an illustration of the differentiable computations and their interdependence.

**a) Obtaining *visibility-aware-weighing*,** $W$: All the modules use a differentiable *visibility-aware-weighing*, $W$ to softly segregate the 3D vertices based on their visibility for a given (or predicted) camera view. The computation of $W$ relies on the fact that visible vertices are influenced by two factors (i) camera and (ii) human skeleton self-occlusion. We identify camera facing vertices using the z-component of the Normal ($N$) while we handle human self-occlusion by soft selection based on a camera-centric depth image and a margin in z-buffering.

$$W^{(k)} \in [0,1] = \exp(-\alpha D^{(k)}) \ \sigma(\gamma N^{(k)}), \text{ where } D^{(k)} = |I^z(v^{(k)}) - Z^{(k)}|$$

**b) Recovering intermediate color,** $\tilde{C}$: Next, we obtain the intermediate, visibility-aware colors, $\tilde{C}$ by weighting the raw picked colors (done by bilinear sampling of image I, given the vertex 2D projection $v^{(k)}$) as shown below.

$$\tilde{C}^{(k)} = I(v^{(k)}) \ (2W^{(k)} - 1), \text{ where } I(v^{(k)}) \text{ denotes RGB color at the } v^{(k)}$$

**c) Applying reflectional symmetry to obtain the full vertex color,** $C$: Next, we focus on propagating the color intensities from the visible vertices (as stored in the intermediate $\tilde{C}$) to the invisible ones (*i.e.* the vertices having low $W^{(k)}$). To realize a fully-colored mesh $C$, we use a predefined, 4-way symmetry grouping knowledge (front-back and left-right) as stored in $S$. First the group colors $\mathcal{C}^{(g)}$ are computed as a normalized combination of the intermediate vertex colors weighted by their visibility weighing $W$. Then, the group colors are directly propagated to all the mesh vertices using $S$ as shown in the following equation.

$$C = S^T * \mathcal{C}, \text{ where } \mathcal{C}^{(g)} = (S^{(g)} \circ \text{ReLU}(\tilde{C}))/(S^{(g)} \circ \text{ReLU}(2W - 1))$$

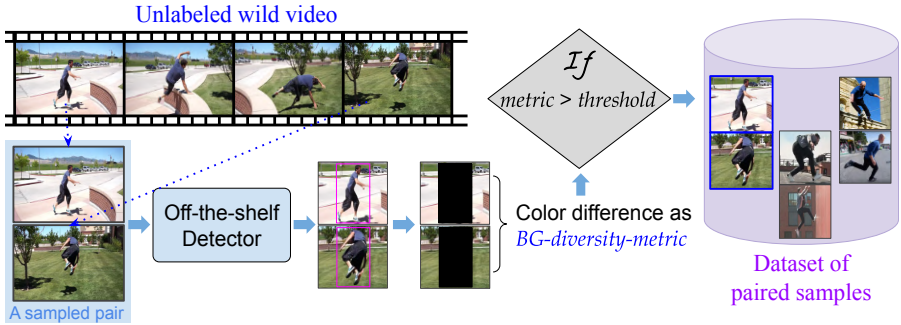

**Fig. 2.** An illustration of the adopted procedure to sample image pairs of diverse background. To built the dataset of image pairs as required by the proposed self-supervised framework, we chose image pairs depicting the same person in diverse pose (maintaining a considerable temporal gap) and background (via *BG-diversity-metric*).

**d) Computation of part-prototype features, $\mathcal{F}$:** Here, we reuse the color-recovery idea to realize part-prototype features. $\mathcal{F}^{(l)}$ is computed as the normalized weighted sum, of the recovered spatial features $\mathcal{H}^{(k)} = H(v^{(k)})$, over the vertices belonging to the part $l$ (using vertex to part-segmentation mapping, $Q$).

$$\mathcal{F}^{(l)} = (\Sigma_{k \in Q^{(l)}} W^{(k)} \mathcal{H}^{(k)}) / (\Sigma_{k \in Q^{(l)}} W^{(k)}), \text{ where } \mathcal{H}^{(k)} = H(v^{(k)})$$

## 2   Sampling image pairs with diverse background

Given a video clip depicting actions of a single person, in consistent apparel, we aim to sample image pairs which would have diverge background (BG) appearance. To realize this, we first prune the video frames using an off-the-shelf person-detector [3] to obtain a reliable human-centric crop as required for the mesh estimation pipeline. Following this, we compute $L2$ distance (mean squared error) between image pairs, only for the regions outside the detector box, to obtain a *BG-diversity-metric* (see Fig. 2). Among all possible frame pairs (beyond 1 sec temporal gap), we choose the pairs having *BG-diversity-metric* greater than a certain threshold value. In contrast to the in-studio datasets with hardly any camera movement implying static BG [1], our in-house collection of YouTube videos have diverse camera movements (*e.g.* Parkour and Free-running videos). The wild camera movement inherently results in huge diversity in the sampled image pairs. Note that, in static camera scenarios BG diversity occurs when the person moves from one location to another. This is because, instead of taking the full video feed, we consider a square region around the detector output as the effective input to the CNN regressor.

## 3   Reflectional symmetry groups

We define reflectional groups where each group constitutes a set of vertices which is assumed to have similar color property. Though this assumption does not hold

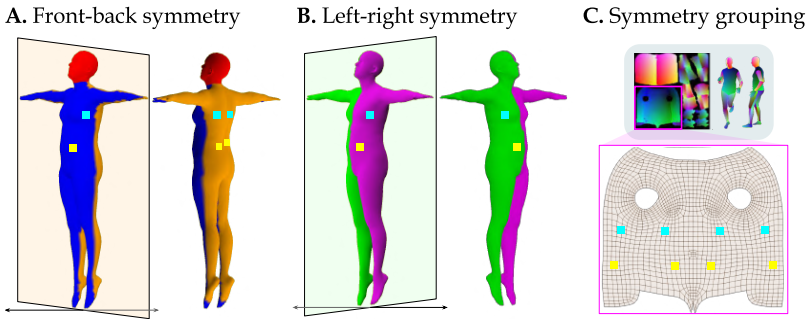

**Fig. 3.** An illustration of front-back and left-right symmetry. This is used to define the multi-hot encoding, $S^{(g)} \in \{0,1\}^K$ which constitutes of four ones indicating vertex members in the symmetry group $g$. Here, "yellow" and "cyan" color patches show rough location of the vertices for two such symmetry groups. Note that, for the head region only left-right symmetry is used (red colored region in panel A).

true in presence of illumination difference and non-symmetric apparel design, we find this to be helpful in general because of the following reasons. Firstly, it is rare to encounter non-symmetric apparel with diverse color difference between the left-right or front-back. Secondly, although the luminosity property (*i.e.* intensity) is influenced in presence of illumination difference, the color property (*i.e.* hue) remains comparable. However, the consistency loss on the part-prototypes and also on the intermediate vertex color $\tilde{C}$ helps us to effectively balance this shortcomings. Broadly, we define 2 types of symmetry groups; a) group sets of 2 members (vertex indices) only for the head region (295 groups), and b) group sets of 4 members (vertex indices) for rest of the body parts (1575 groups). See Fig. 3 for a rough illustration. 4-membered groups are obtained by applying both front-back and left-right symmetry. However, 2-membered groups represent only left-right symmetry. Note that all the group sets are mutually exclusive and exhaustive, *i.e.* 2*295 + 4*1575 = 6890, where 6890 is the total number of vertices. This symmetry knowledge is stored as a multi-hot encoding denoted as $S^{(g)} \in \{0,1\}^K$ which constitutes of four ones indicating vertex members in the symmetry group $g$. All the symmetry groups are combined in a symmetry-encoding matrix represented as $S \in \{0,1\}^{G \times K}$. This multi-hot symmetry group representation helps us to perform a fully-differentiable vertex color assignment for all the vertices including the occluded and non-camera facing ones.

## 4    Qualitative evaluation

In order to evaluate the generalizability of our model, we visualize our model's 3D pose and shape performance on a variety of images sampled from different datasets. Fig 4 shows the predicted colored mesh and the corresponding 3D pose in aligned grid plots. Fig. 5 shows a qualitative analysis on the standard 6-part mesh overlay. Here, mesh overlay can be considered as a proxy to evaluate both shape and pose in a collective fashion.

**A**. Results on YouTube, LSP and 3DPW dataset (in-the-wild)

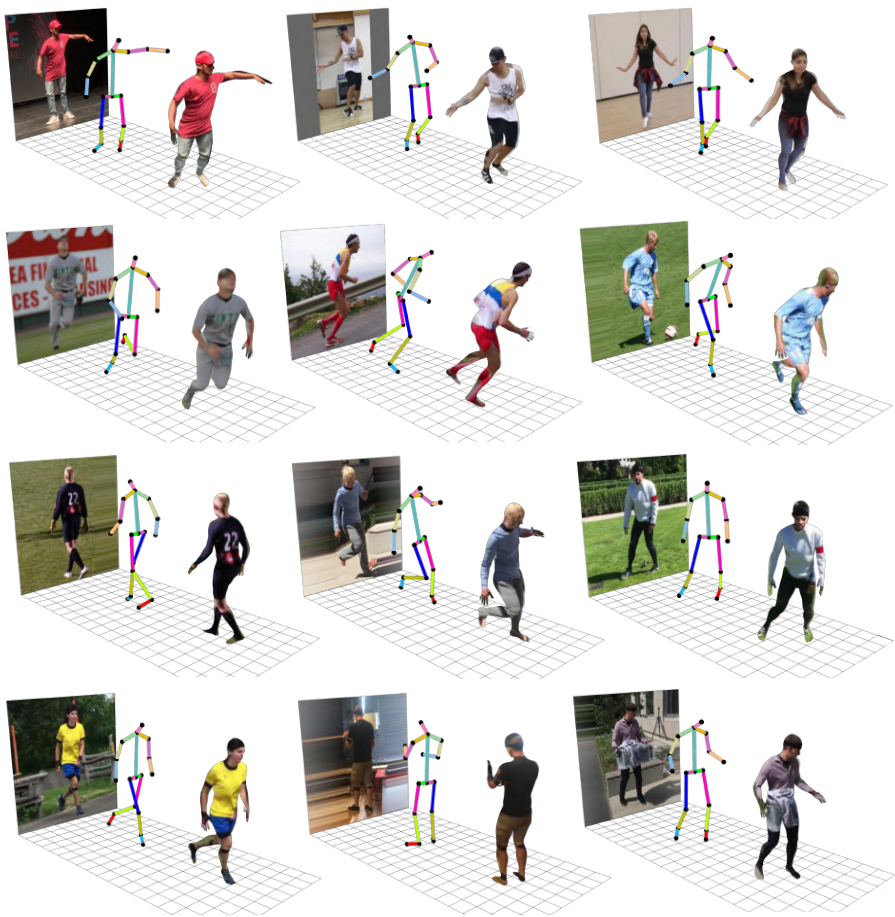

**Fig. 4.** Qualitative results on single image colored human mesh recovery.

Although, pose and shape are predicted correctly, background leaks could occur at misaligned locations that can be visualized using the coloured mesh reconstructions as shown in Fig 4. Also note that, SMPL [2] does not parameterize hand pose hence hands remain in a fixed mean pose (flat open hand). This tends to be a consistent location for background leakage. Background leakages are observed to generally occur at boundaries of hands and feet; *e.g.*, in row 2 last column of Fig 4 the green background leaks onto the appearance of the hand due to the limitation of the parametric human model in articulating the exact hand pose. Also, our model outputs sub-optimal results in cases with complex inter-limb occlusions as highlighted in magenta in Fig. 5.

**A**. Results on H36M dataset (in-studio)

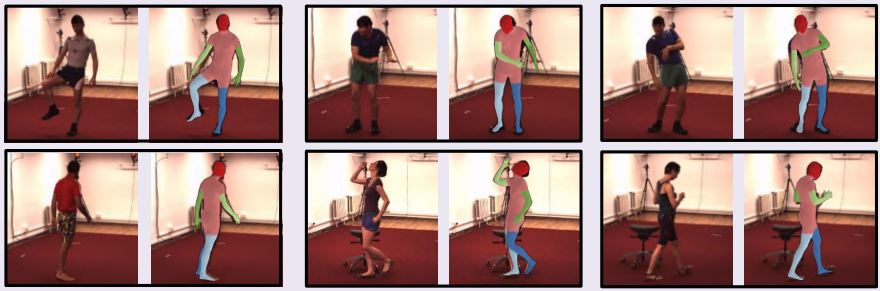

**B**. Results on 3DPW dataset (in-the-wild)

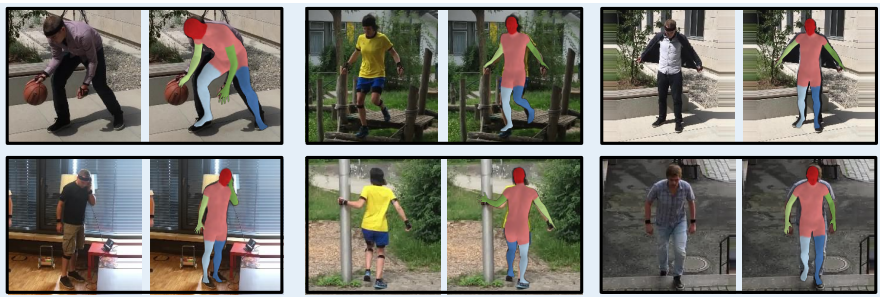

**C**. Results on LSP dataset (in-the-wild)

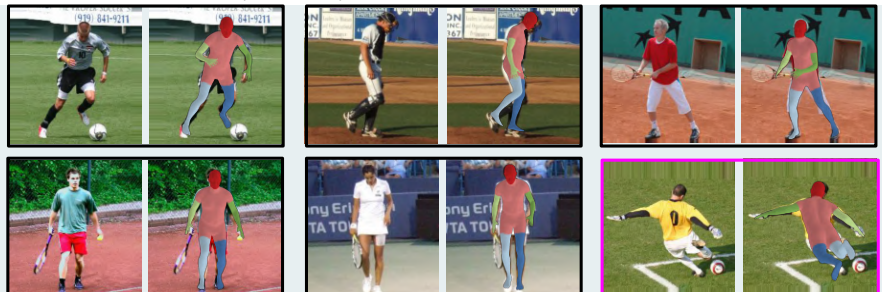

**D**. Results on YouTube dataset (in-the-wild)

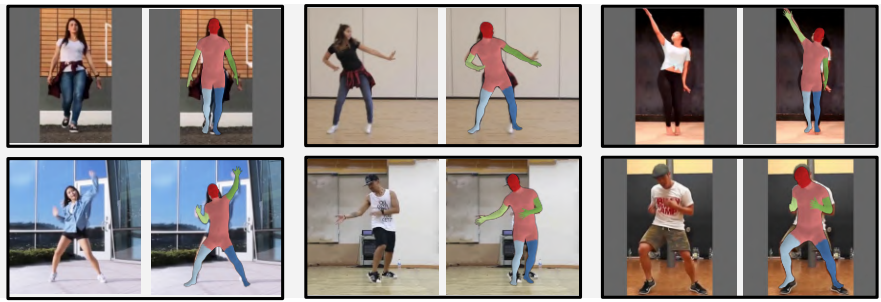

**Fig. 5.** Qualitative results. In each panel, 1st column depicts the input image, 2nd column shows the model-based part segments on **A.** Human3.6M (in-studio), **B.** 3DPW (in-the-wild) **C.** LSP (in-the-wild) **D.** YouTube (in-the-wild). The model fails in presence of complex inter-limb occlusions (in magenta box).



\bibliographystyle{splncs04}
\bibliography{egbib}

\end{document}